\documentclass[sigconf]{acmart}

\usepackage{booktabs,tabularx,stfloats,microtype,enumitem} 
\usepackage{float}
\AtBeginDocument{%
  \providecommand\BibTeX{{%
    \normalfont B\kern-0.5em{\scshape i\kern-0.25em b}\kern-0.8em\TeX}}}

\copyrightyear{2022} 
\acmYear{2022} 
\setcopyright{acmcopyright}
\acmConference[IDC '22]{Interaction Design and Children}{June 27--30, 2022}{Braga, Portugal}
\acmBooktitle{Interaction Design and Children (IDC '22), June 27--30, 2022, Braga, Portugal}
\acmPrice{15.00}
\acmDOI{10.1145/3501712.3529721}
\acmISBN{978-1-4503-9197-9/22/06}

\acmSubmissionID{9702}

\begin{document}

\title{Exploring Children's Preferences for Taking Care of a Social Robot}

\author{Bengisu Cagiltay$^*$, Joseph Michaelis$^{**}$, Sarah Sebo$^{***}$, and Bilge Mutlu$^*$}
\affiliation{%
  \institution{$^*$Department of Computer Sciences, University of Wisconsin--Madison}
  \country{Madison, Wisconsin, USA}}
\affiliation{%
  \institution{$^{**}$Learning Sciences Research Institute, University of Illinois Chicago}
  \country{Chicago, Illinois, USA}}
\affiliation{%
  \institution{$^{***}$Department of Computer Science, University of Chicago}
  \country{Chicago, Illinois, USA}}
\email{bengisu@cs.wisc.edu, jmich@uic.edu, sarahsebo@uchicago.edu, bilge@cs.wisc.edu}



\renewcommand{\shortauthors}{Bengisu Cagiltay, Joseph Michaelis, Sarah Sebo, and Bilge Mutlu}


\begin{abstract}

Research in child-robot interactions suggests that engaging in ``care-taking'' of a social robot, such as tucking the robot in at night, can strengthen relationships formed between children and robots. In this work, we aim to better understand and explore the design space of caretaking activities with 10 children, aged 8--12 from eight families, involving an exploratory design session followed by a preliminary feasibility testing of robot caretaking activities. The design sessions provided insight into children's current caretaking tasks, how they would take care of a social robot, and how these new caretaking activities could be integrated into their daily routines. The feasibility study tested two different types of robot caretaking tasks, which we call \textit{connection} and \textit{utility}, and measured their short term effects on children's perceptions of and closeness to the social robot. We discuss the themes and present interaction design guidelines of robot caretaking activities for children.
\end{abstract}



\maketitle
\section{Introduction}

Social robots are used in a range of domains including education, healthcare, entertainment \cite{belpaeme2018social, breazeal2011social, correia2017social} and in environments such as schools, hospitals, homes, workplaces, public spaces \cite{komatsubara2014can, jeong2015social, scassellati2018improving, lee2012ripple,tonkin2018design}. Prior research has explored how robots might assist children in these domains and environments in a variety of ways, including helping children develop physical exercise skills \cite{guneysu2017socially}, learn a second language \cite{vogt2017child}, manage anxiety in healthcare settings \cite{jeong2015social}, and form social connections \cite{bjorling2020can}. Many of these applications will require robots to establish and maintain lasting relationships with children. 
Although research on long-term child-robot interaction \cite{scassellati2018improving, michaelis2018reading, de2016long, clabaugh2018month, randall2019more} is growing, designing for long-term use remains an open challenge in human-robot interaction design. In particular, what design elements or activities might facilitate the forming of a bond between a child and a robot remains underexplored.
Design features or interaction strategies that facilitate short-term interactions may not be effective in helping the robot establish long-term relationships. Furthermore, a robot that is embedded in a family environment and that ``lives'' with the child might be designed to engage in activities that are unique to the intimacy, routines, and expectations of such a setting. In this paper, we explore the activity of \textit{caretaking} of a social robot as a new and unique responsibility introduced by in-home, long-term interactions with a robot. Caretaking might range from maintaining the robot by charging and cleaning, protecting it by providing a shelter and a safe space, or providing companionship by chatting, playing games, or reading together. It also serves as a new and exciting design space with the potential to craft a positive user experience that facilitates the forming of lasting bond between the child and the robot.
\begin{figure*}[t!]
    \centering
    \vspace{-6pt}
    \includegraphics[width=0.8\linewidth]{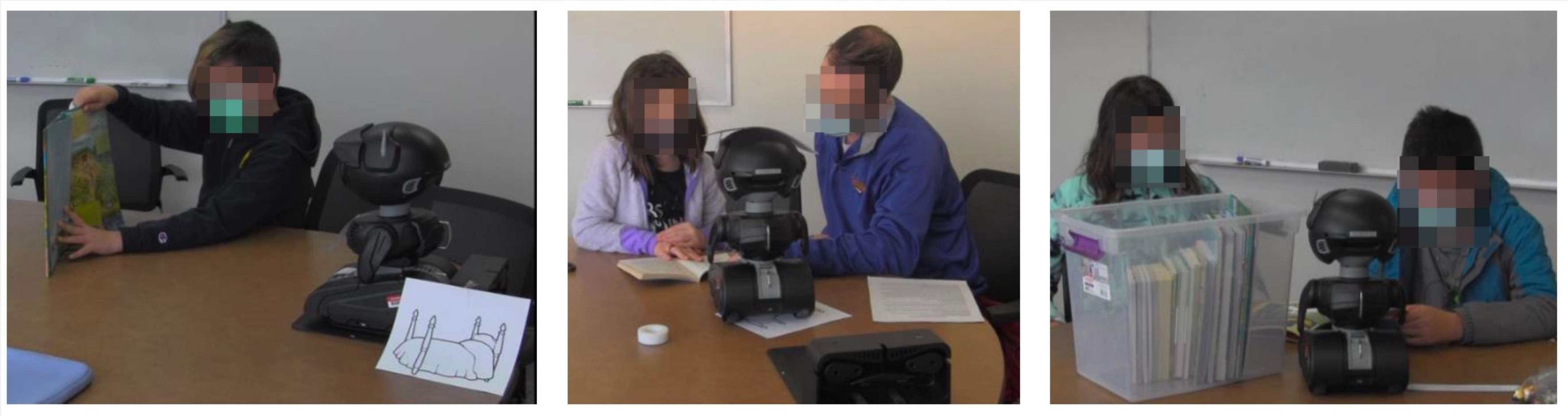}
    \vspace{-6pt}
    \caption{\textit{Children reading to the robot for a bedtime care routine}: In this study we explored children's preferences for taking care of a social robot and conducted a feasibility study to evaluate children's experience of taking care of a robot. Our findings provide insight into designing social robot caretaking activities for children.}
    \label{fig:my_label}
    \vspace{-6pt}
\end{figure*}


Similar to a friendship between two people or the connection between a person and a real or artificial pet (\textit{e.g.,} the Tamagotchi in a key-chain form\cite{marti2005engaging}), we aim to explore how meaningful, lasting interpersonal connections might be formed between children and robots by designing a robot capable of also \textit{receiving} care \cite{tanaka2010care} rather than simply giving care. Prior research has shown that caring for an interactive toy, robot, or chatbot by comforting \cite{lee2019caring}, teaching \cite{matsuzoe2014learning}, or touching \cite{shiomi2017does} it can help form stronger connections and facilitate positive outcomes such as improvements in mental health in adults \cite{lee2019caring}, learning gains in children \cite{matsuzoe2014learning}, and the amount of time adults spent on a monotonous task \cite{shiomi2017does}. Building on these findings, we posit that the key to developing a social companion robot for families, particularly for children, that overcomes barriers to successful long-term human-robot interaction is the formation of lasting interpersonal connections between the robot and its users. We explore how designing a robot that is capable of \textit{receiving} care from its users might facilitate the forming of such relationships. Our work aims to address the following questions:

\begin{enumerate}[leftmargin=0.5in]
    \item[\textbf{RQ1.}] What types of in-home robot caretaking tasks will be performed regularly by children?
    \item[\textbf{RQ2.}] How can we design robot caretaking tasks with children to foster interpersonal connections and long-term use of the robot?
\end{enumerate}

As a first step toward addressing these questions, we conducted an exploratory design study that explored what types of caretaking tasks children would prefer and how these care obligations would fit in their daily routines. We then conducted a brief feasibility study that tested two types of caretaking tasks and measured how children perceived the robot and how close they felt to the robot after completing the care tasks. This work makes the following contributions.

\begin{enumerate}[leftmargin=0.3in]
\item A characterization of \textit{caretaking} as a design space for child-robot interaction design;
\item An understanding of children's preferences for robot caretaking activities;
\item A preliminary understanding of children's perceptions of designed caretaking interactions.
\end{enumerate}

\section{Background}

\paragraph{Caretaking in Child-Robot Interactions} Limited work shows how people can benefit from expressing care towards non-robot applications, such as chatbots. For example, people whom interacted with a care-receiving chatbot by comforting it had increased positive mental health outcomes in self-compassion \cite{lee2019caring}. In another study \cite{vsabanovic2014designing}, a social robot deployed in the workspace for 10 work days was found more effective in helping users better manage their breaks. The social robot responded to ``petting,'' displayed agitation for successive break alerts, and needed to be ``fed'' fruits represented by RFID cards.
Research focused on caretaking in child-robot interactions is limited, however some studies reported children spontaneously demonstrating robot care behaviors. For example, a social robot was deployed in an early childhood education center for five months, and toddlers showed more caretaking behaviors for the robot (such as putting a blanket on the robot while saying ``night-night'') than other objects such as an immobile robotic toy, treating the care-receiving robot as a peer \cite{tanaka2007socialization}. As part of their findings, \citet{tanaka2007socialization} reported a set of factors that impacted the quality of children's interaction with the robot, including caretaking, rough-housing, hugging, touching the robot with objects, or touching the robot's head, face, trunk, arm, hand, leg, or foot. These factors manifested in children's behaviors differently depending on the robot, for example, violent rough-housing was observed only for the immobile robotic toy, while children frequently hugged the care-receiving robot. \citet{tanaka2010care} explains that such care activities are a part of children's development of \textit{learning by teaching}. They argued that, the robot being a ``weaker'' entity than the children, in size and sophistication, motivated children's care interactions with the robot in the classroom. However, research exploring caretaking in child-robot interactions are typically limited to a classroom context and the literature lacks insight into how robot caretaking might be formed in \textit{children's homes}. While our work is also situated in a laboratory context, we aim to extract initial design factors for children's robot caretaking tasks that would later be transferable to the home setting.



\paragraph{Long Term Child-Robot Interactions} The design of social robots that can sustain long-term interactions with children is key for exploring how children would take care of a robot in their homes. However, human-robot interaction (HRI) researchers \cite{fernaeus2010you,de2017they,weiss_merely_2021} as well as companies that have developed social robots (\textit{e.g.}, Jibo, Anki, Kuri) have continued to face barriers to sustaining long-term human-robot interactions. While social robots are engaging for the first few days or weeks, due to the ``novelty-effect'', researchers find that a majority of human subjects stop interacting with social robots altogether before the end of six months \cite{de2017they, fernaeus2010you, weiss_merely_2021} and the companies that have developed social robots for in-home use have faced limited sales and discontinued products \cite{hoffman_anki_2019}. People who stop interacting with social robots report disappointment in the robot’s capabilities \cite{fernaeus2010you, leite2009time, sung2009robots}, annoyance with verbal interjections by the robot \cite{de2017they}, and difficulty finding a purpose for the robot \cite{caudwell_irrelevance_2019, weiss_merely_2021}. HRI researchers and social robot developers in industry have tried to address these shortcomings by adding new content or capabilities to the robot to prevent interactions with the robot from becoming stale \cite{leite2013social, tanaka2007socialization}, developing adaptive algorithms that seek to personalize the robot's interactions to each user \cite{gordon2016affective, park2019model, ramachandran2019personalized}, and increasing the robot's use of emotion and sense of personality \cite{kirby2010affective, leite2013social, weiss_merely_2021}. However, despite advances in each of these areas, social robots still manage to fail sustained long-term interactions with people, demonstrating that the field still lacks a clear understanding of how to design social robot interactions to sustain long-term use. We aim to contribute to the HRI field by providing an understanding of how children would take care of a social robot at home and exploring how caretaking can help form \textit{stronger connections between children and robots}. We believe these stronger connections can contribute to supporting sustained long-term interactions with robots, and within our current work, we aim to explore the factors that could contribute to long-term use through the design of robot caretaking tasks for children.


\section{Method}

To explore and evaluate the feasibility of robot caretaking tasks a child might do at home, we conducted a study consisting of two parts: (1) interviews and family discussions related to caretaking tasks at home, and (2) a child-robot interaction feasibility study testing children's preferences towards two types of caretaking tasks. The study was conducted in a university laboratory in-person visit adhering to campus COVID-19 guidelines and lasted 1--1.5 hours. 

The first author facilitated each study session. We started by describing the study to the parent and child and provided a written consent form to the parent. After obtaining informed consent from the parent, we addressed the participating children to further describe the goal of the study and the structure of the session and respond to their questions. The study was initiated after children explicitly shared verbal assent to participate in the study.

\subsection{Part 1. Family Discussions} 
The family discussions were structured to understand current caretaking tasks and chores children did at home, how these tasks could extend to a social robot, and how these tasks would fit into daily routines (such as morning, nighttime, and playtime routines) of children and family members. During these sessions both children and parents were involved in the prompted activities together and were encouraged to discuss their answers with each other.

\textit{Current caretaking tasks.} 
We first explored how current caretaking tasks are formed at home by conducting a brief interview and discussion session with the participating children and parents. We asked the child and the parent, \textit{``what are the current chores, tasks, and responsibilities you/your child take care of at home,''} and took notes of each care task mentioned by the participants along with how often each care task was conducted, and by whom.

\textit{Taking care of a social robot.}
After discussing children's current caretaking tasks, we shared a definition of a ``social robot'' and presented children with a powered down social robot to give them a sense of the size, shape, and features of the robot without overly biasing their expectations about the robot's functions. After introducing the robot we described its capabilities including being able to move around, see its surroundings, speak, make facial expressions, and move its arms. We then reminded the children and parents of each caretaking task and chore shared from their earlier discussion by going over the list together, and asked if and how these tasks and chores could be transferred to taking care of a social robot. We took note of new robot caretaking tasks and prompted the children and parents for further discussion and brainstorming. 

\textit{Taking care of a social robot in daily routines (morning, nighttime, playtime).} After discussing social robot caretaking tasks, we introduced three routines (morning, nighttime, and playtime) and asked the child how their routines would be shaped by including the previously discussed robot caretaking tasks, and took notes of the steps the child described for each routine. Once completed, we asked the child to rank each routine based on how likely it is for them to do the routine, how much they would enjoy the routine, and how much help they would need from a parent/adult. Finally, we asked the child to rank all three routines from the one they liked the most to the one they liked the least and describe their reasoning after each ranking. These rankings were mainly used to highlight any potential contrasts between preferences of robot care routines and to promote further discussion to better understand children's reasoning for preferring particular caretaking tasks over others.  


\subsection{Part 2. Feasibility Study}
After completing the interviews, we conducted a brief feasibility study to explore the short-term effects of robot caretaking tasks on children's experiences with and perceptions of the robot. We randomly assigned families to one of two conditions: ``connection'' and ``utility.'' Both conditions involved the child \textit{taking care of the robot}, but they varied in the form of caretaking. In both conditions, children read to the robot (named Micky) for a bedtime routine, but depending on their assigned condition, they followed different tasks before and after reading to the robot. The \textit{connection} caretaking tasks focused on taking care of the robot by directly interacting with it, while the \textit{utility} caretaking tasks focused on taking care of chores in the robot's surroundings and involved indirect interaction with the robot.

\textit{Connection} tasks focused on caretaking activities that emphasized social connections and closeness between the robot and child. Activities were selected based on high levels of touch and care directed at the robot. Participants were provided with a task sheet, a box of books as a bookshelf, and paper cutouts of pajamas, a blanket, and bed sheets for the robot. The task sheet had three written instructions: 
(1) put Micky’s pajamas on and prepare the bed;
(2) pick a book and read it to Micky until Micky falls asleep;
(3) once asleep, hug Micky, place Micky on the charging pad, and use the blanket to tuck Micky into bed.

\textit{Utility} tasks focused on caretaking tasks that were useful or helpful to the robot. Activities included chore-like tasks that involved cleaning and organizing the environment of the robot. Participants were provided with a task sheet, a box of books as a bookshelf, laundry basket, and paper cutouts of toys and clothes for the robot.
The task sheet had three written instructions: 
(1) organize Micky's bookshelf from largest to smallest book;
(2) pick a book and read it to Micky until Micky falls asleep;
(3) once Micky is asleep, clean up Micky's toys and take Micky's dirty clothes to the laundry.

After providing the instructions, the facilitator started the video recording, left the room, and remotely controlled the robot's expressions such that the robot displayed a ``sleepy face'' 10 minutes after the facilitator left the study room to indicate that the robot fell asleep. This 10-minute period allowed participants to complete their first task and start reading to the robot as part of their second task. Children started their third task only after seeing the robot's sleepy face. After completing all three tasks, the child left the study room and informed the facilitator that they were done.

\paragraph{Measures}
We asked children to respond to two previously validated questionnaires:\footnote{The questionnaires used in this study are available as open-access resources \url{https://osf.io/dj7hb/?view_only=b4285bfd321841aa93705c7010871ae1}} (1) robot perceptions scale \cite{parise1999cooperating, mutlu2006task} and (2) robot closeness scale \cite{gachter2015measuring, van2020closeness}. The \textit{robot perceptions scale} included 14 items, rated on a seven-point scale, with questions asking about the child's opinion of their robot partner. The questions were grouped under five factors: mutual liking, attractiveness, humanlikeness, sociability, and intelligence of the robot (e.g., ``How much did you like your partner;'' ``How bored or excited was your partner''). The \textit{robot closeness scale} included six items rated on a five-point scale. Children responded to items that demonstrated increasingly overlapping circles and asking to select the one that ``best shows your relationship with the robot'' \cite{gachter2015measuring} and a five-item scale of closeness that asked children's opinions of the robot, Micky (e.g.,``Micky feels like a friend to me;'' ``Micky and I are a good match'') \cite{van2020closeness}. We then asked semi-structured interview questions to understand the family members' experiences during the caretaking task. Finally, we provided the parent with a demographics survey and concluded the study. The feasibility study was video recorded. 

\begin{figure*}[t!]
    \centering
    \vspace{-6pt}
    \includegraphics[width=0.8\linewidth]{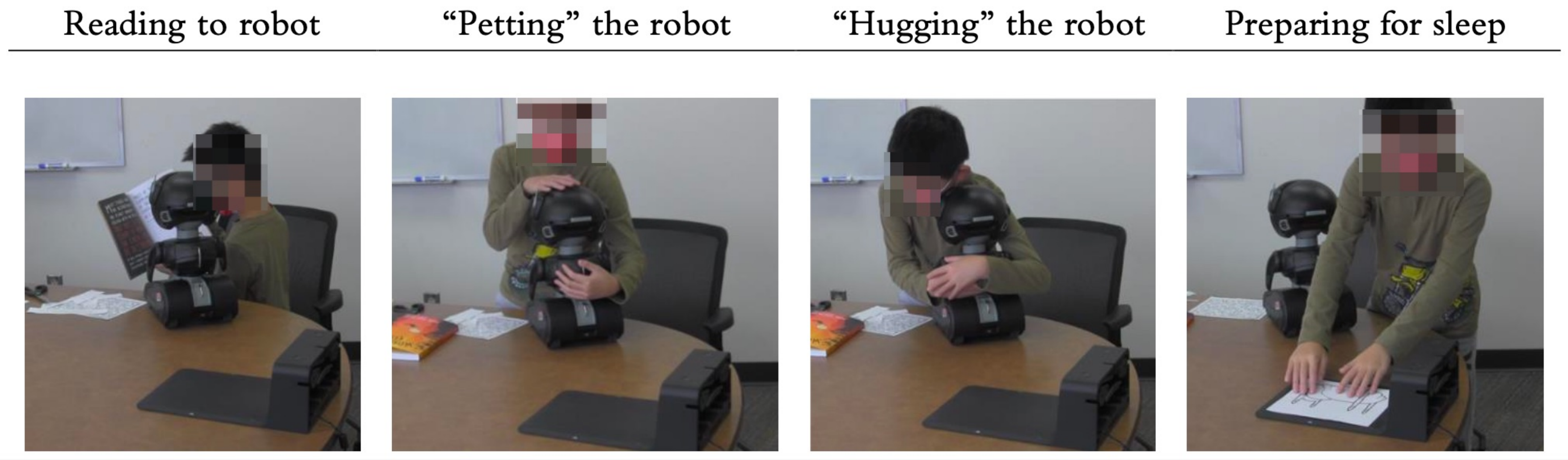}
    \vspace{-12pt}
    \caption{\textit{Feasibility Study (Connection Condition):} Children prepared the robot's bed by attaching paper cutouts of pajamas on the robot, read to the robot until it was asleep, then hug the robot and place it on a charging pad as a bed for the robot to sleep.}
    \label{fig:my_label}
    \vspace{-6pt}
\end{figure*}



\subsection{Materials} 
We used Misty II\footnote{\url{https://www.mistyrobotics.com/}} as a social robot platform throughout the study due to its capability to express emotion, physical humanoid form, size, and mobility that allows children to easily interact with. Additional resources provided during the family discussions included arts and crafts supplies to support pretend play. For the feasibility study, we provided a set of age-appropriate books and paper cut-outs of resources (e.g., toys, pajamas, bed sheets, etc.) needed in each condition.

\subsection{Participants}
We recruited families with children aged 8--12 through email solicitation where eligible and non-eligible siblings were allowed to participate. Eleven children from eight families participated and ten children were in the eligible age range (seven boys, three girls, $M=10.1, SD=1.19$). For the feasibility study, five children (four boys, one girl) were assigned to the \textit{utility} condition, and five children (three boys, two girls) were assigned to the \textit{connection} condition. Families were randomly assigned to each condition. Each family received \$15 compensation after completing the study.

\subsection{Analysis} 
We conducted a thematic analysis on the transcriptions of family discussions and semi-structured interviews. We followed the analysis guidelines by \citet{braun2019thematic} and \citet{mcdonald2019reliability} where the first author was familiarized with the data through conducting the study sessions, transcribing the interviews, and reviewing the video recordings. The author coded the data and discussed the candidate themes with the study team members which were reviewed, refined, and reported as themes. We conducted an exploratory statistical analysis on the survey response findings, including independent samples t-tests to compare the survey responses based on the utility and connection conditions.



\section{Results}
\subsection{Exploring Robot Caretaking Tasks}

\paragraph{Current caretaking tasks and chores}
During this part of our family discussion, the theme that emerged around children's current caretaking tasks suggested their responsibilities typically involved care for pets and younger siblings, and chores that involved cleaning their spaces and helping with family chores. Most commonly discussed \textit{caretaking tasks} were feeding/playing/cuddling pets, paying attention and giving love to pets, babysitting younger siblings, and most commonly discussed \textit{chores} were tidying up rooms/toys/bookshelves/clothes, making the bed, helping with dinner/laundry/dishes/garbage disposal. Children with siblings mentioned they would take turns in many of their tasks and chores, especially ones related to taking care of a pet.  

\paragraph{Taking care of a social robot}
The theme that emerged from our discussion of caring for a social robot suggested that children most commonly associated tasks that took care of the robot's basic needs, while also considering some socially demanding tasks as part of robot caretaking responsibilities. The most commonly discussed \textit{robot caretaking tasks} imagined by families involved feeding, cleaning, protecting, offering affection and teaching the robot. Specifically families mentioned feeding the robot by charging it, and cleaning the robot by dusting or wiping the robot, preparing the robot's bed or cleaning its sheets. Children also suggested protecting the robot from crashing or falling down stairs, making sure it doesn't overheat, and protecting it from other pets. They also suggested showing affection with helping the robot fall asleep or wake up, and making sure the robot is treated kindly and gets attention and love. Finally, some families mentioned caretaking tasks could include playing and teaching games to the robot, or teaching the robot to get along with other pets and to be nice to people. 





\paragraph{Taking care of a social robot in daily routines}
Children discussed how they would include robots into their daily routines (i.e., morning, nighttime, playtime), creating a personalized \textit{robot care routine}. The robot care activities children included in their \textit{morning routines} mostly focused on waking the robot up and making sure the robot is ready for the day, feeding/charging the robot, cleaning/dusting the robot, morning exercise with the robot, and playing/chatting/reading with the robot during the day. One child 
expressed they would charge the robot during breakfast so they could all be having ``breakfast together as a family.'' In their \textit{nighttime routines}, children discussed care obligations such as feeding/charging the robot, cleaning/dusting the robot, prepare the robot's bed and putting the robot into bed, and share bedtime activities such as reading/playing/watching movies/listening to music together with the robot. Nearly half of the children suggested the robot would have a ``protective box'' or ``safe blanket'' that would be used as a bed for the robot where it would be kept clean, protected from other pets, or protected by things the robot is scared of, like thunderstorms at night. 
As a part of their \textit{play routines}, typically on weekends, children generally discussed carrying the robot to the playroom, preparing and tidying up the play area, teaching or playing a game with the robot.  



Discussions about personalized robot care routines revealed that children are generally willing to take care of the social robot when needed, however, many children admitted that it might be challenging to maintain the caretaking tasks without having a routine. Two children mentioned they would need help from their parents to remember taking care of the robot or to carry the robot up/down stairs. Some children preferred taking care of the robot in their morning routines over nighttime routines, stating that ``nighttime is my time to relax and get some sleep,'' while other children shared opposite opinions expressing that mornings might be difficult because they ``need to get ready for school.'' 


\subsection{Feasibility Evaluation of Robot Caretaking}
\subsubsection{Video Analysis Results}
All participating families successfully completed the robot caretaking tasks assigned to them. Two families that participated in the utility condition with eligible siblings were observed to take turns with each other in completing their tasks (e.g., to sort the books or read to the robot),  
pointed out by one of the parents saying \textit{``you both are so collaborative!''} Parents in general were indirectly involved in the caretaking tasks, ranging from waving or talking to the robot, repeating the tasks, suggesting ways to complete the task, responding to children's questions, and motivating the child to complete the tasks. For example, one child in the connection condition  
initially laid the robot on its back on the bed, and the parent expressed \textit{``Micky sleeps differently than you sleep,''} prompting the child to change the robot's placement. 
Another child's parent motivated them by saying \textit{``it seems like you're doing a nice job taking care of the robot!''} Two children from the connection condition 
\textit{petted} the robot's head after completing their tasks, and one child 
said \textit{``good night''} after the robot fell asleep, but similar attempts of contact or interactions with the robot was not observed in the utility condition. All children in the connection condition were observed to talk by whispering after the robot fell asleep, while children in the utility condition did not display similar behaviors. 


\subsubsection{Survey Results} 

\paragraph{Robot Perceptions}
The measure included five categories, humanlike, attractiveness, sociable, intelligence, and mutual-liking. There were no significant differences between the children's perceptions of the robot's \textit{human-likeness} $(t(8)=1.96, p=.085)$, \textit{attractiveness} $(t(8)=.86, p=.41)$, and \textit{mutual-liking} $(t(8)=1.4, p=.19)$. However, the robot's perceived \textit{sociability} ($t(8)=3.42, p=.009$) and \textit{intelligence} ($t(8)=2.82, p=.022$) was significantly different. Children in the utility connection perceived the robot as more sociable ($M=6.04, SD=.83$) and intelligent ($M=6.1, SD=.76$) compared to the children in the connection condition (sociable $M=4.4, SD=.67$; intelligent $M=4.75, SD=.75$). Overall, these exploratory findings suggest that children perceive a robot more \textit{sociable} and \textit{intelligent} when conducting caretaking tasks that are aimed to be useful or helpful to the robot, compared to conducting tasks that aim to form social connections and closeness.

\paragraph{Robot Closeness}
The measure included an item with overlapping circles indicating relationship level and five items indicating closeness. The \textit{relationship circle} item revealed that children in the connection condition ($M= 4, SD=.7$) felt significantly closer to the robot ($t(8)=2.74, p=.025$), compared to the children in the utility condition ($M=2.6, SD=.89$). The collection of the five \textit{closeness} items only revealed marginal significance in perceived closeness ($t(8)=2.14, p=.051$) between the connection ($M= 4.12, SD=.62$) and utility condition ($M= 3.48, SD=.22$). Additional tests on each independent item for the closeness scale revealed that children perceived significantly higher levels of \textit{comfort} with the robot ($t(8)=2.13, p=.011$) and more strongly felt that they were \textit{becoming friends} with the robot ($t(8)=2.13, p=.011$) in the connection condition (comfort, $M=5, SD=.0$; becoming friends, $M=3.8, SD=.83$) compared to children in the utility condition (comfort, $M=4.2, SD=.83$; becoming friends, $M=3, SD=.0$). In sum, these exploratory findings infer that children tend to feel more comfortable and as if they are forming friendships when conducting caretaking tasks designed to form social connections and closeness.

\section{Discussion and Conclusion}
We explored caretaking tasks that children might engage in with a social robot and how those tasks might affect their perceptions of the robot. We found that children currently engage in caretaking activities with pets and younger siblings and children would want to engage in caretaking activities in which they feed, protect, teach, and show affection toward a social robot. Our initial exploration revealed some evidence that caretaking activities that focus on connection-making with the robot better promote a sense of closeness, comfort, and friendship, whereas utility activities promote a sense that the robot is sociable and intelligent, possible because the robot is seen as an independent agent that does not need caretaking \cite{mutlu2021virtual}. These findings suggest that caretaking activities may serve as an effective mechanism to promote engagement with social robots and provide early evidence of the types of activities that may be most conducive to connection making between a child and a robot. In this section, we explore the nuances of these findings and their relation to prior work, present design implications based on these findings, and describe future work needed to address the limitations of the current preliminary study.

\paragraph{Caretaking Tasks for Child-Robot Interactions} Exploratory design sessions showed how children and their families imagine engaging in caretaking tasks with a social robot. Of note, we found families to describe ways to take care of the robot through activities that simulate feeding, protecting, teaching, and showing affection to the robot. Different families indicated that whether they would engage in an activity would depend on the time of day and that their children currently took care of younger siblings and pets. Consistent with prior work \cite{vsabanovic2014designing}, we observed participants to simulate feeding the robot through charging and showing affection to the robot through petting.


\paragraph{Impact of Caretaking Tasks on Perceptions of the Robot} Our exploratory findings indicate that children's perceptions of the robot differed across caretaking tasks. While we acknowledge that more work is needed to understand underlying reasons for these differences, our preliminary findings motivate further exploration of the design space of robot caretaking tasks and understanding of their effects on children's perceptions of and closeness toward robots.

\subsection{Design Implications}
In general, children expressed that they enjoyed the activity of taking care of a social robot, regardless of the condition they were assigned to. While children in this age range were able to complete all caretaking tasks, it is possible that younger children might find some tasks more challenging, and older children might find them trivial. Thus, robot caretaking tasks should consider the child's age, integrating a set of age-appropriate caretaking tasks for primary users and siblings in the household. Furthermore, families might benefit from robot caretaking tasks that would allow them to partake in a shared activity, while allowing the child to lead the robot care and require minimal help from parents. 

The \textit{personalized robot care routines} showed that different children might prefer taking care of the robot at different times of the day. To support sustained caretaking from children, robot caretaking tasks should be designed to adapt to the child's routine, e.g., by learning their preferences over time. For example, if a child avoids taking care of the robot at night, the robot might prompt the child to engage in caretaking activities at alternatives times of the day through verbal or nonverbal expressions or through explicit requests 
(e.g., wanting to \textit{``go to bed,''} needing to be fed/charged, asking for a bedtime story). Similarly, in a household where children have trouble getting ready for school in the morning, over time the robot might reduce its requests in the morning and identify an alternative time of the day for care requests. 



\subsection{Limitations and Future Work}
Our work has a number of limitations. Firstly, due to the COVID-19 pandemic, it was challenging to recruit in-person participants, as the timeline of this study overlapped with steep increases in cases within our target demographic. It was also challenging to maintain a balanced gender distribution and increase the sample size within our recruitment efforts. We acknowledge that our findings are preliminary and exploratory, and a study with a larger sample size is necessary to draw generalizable conclusions. However, we believe that our design guidelines for future work can serve as the initial groundwork for designing robot caretaking tasks for children. 

The short-term interaction involved in our study limits our ability to generalize findings to long-term interactions. Our study only explored possible robot caretaking scenarios and their short term effects on children's robot perception and connection, and long-term studies are necessary to understand how such care obligations can form sustained interactions between children and robots. Future research should prototype autonomous support for robot caretaking tasks to allow long-term field testing. Furthermore, children's routines might also be shaped by external factors that were not a focus of our exploratory study, for example parents' routines, which should be taken into consideration when designing for in-home settings. Future prototypes can also integrate additional social features, such as verbal and non-verbal expressions and affective responses for communication during caretaking. Some example scenarios might include the robot (1) expressing ``sleepiness'' around bedtime; (2) expressing ``disgust'' when it has been a long time since its last ``bath/cleaning;'' (3) expressing ``jitters'' around the time of day when it usually is taken for a walk/exercise; (4) expressing its ``worries'' when it encounters a fear and asks for a hug and reassurance; and (5) the robot saying \textit{``thank you, I really like it when you take care of me so well.''} These features will help the robot communicate to the child when to care activities might be started and as serve as positive feedback to the child when they take care of the robot.

In conclusion, through our exploratory design sessions, children shared their current caretaking tasks and expressed their thoughts and preferences for robot caretaking tasks and activities they would do at home. Our feasibility study investigated how children experience different types of robot caretaking tasks, and the exploratory findings revealed that children might perceive the robot differently based on the type of caretaking task they partake in. Overall, our findings serve as a basis for designing robot caretaking tasks for children. 

\section{Selection and Participation of Children}
This study protocol was reviewed and approved by the University of Wisconsin-Madison Institutional Review Board (IRB). Children aged 8--12 were recruited through their parents who were contacted through university mailing lists. During the study, researchers described the procedure, encouraged children and parents to ask questions, obtained written consent from parents, and verbal assent from minor(s). Participants were informed that they will be audio and video recorded during the study but their confidentiality will be protected through proper anonymization. We also described that the study is not a test and there are no right or wrong answers to any of the questions. The study was initiated after the minor(s) clearly and verbally consented to participate. Parents received \$15 compensation for their time.

\begin{acks}
This work was supported by NSF DRL-1906854: ``Designing and Testing Companion Robots to Support Informal, In-home STEM Learning.''
\end{acks}

\bibliographystyle{ACM-Reference-Format}
\bibliography{sample-base}

 \end{document}